\documentclass[11pt,a4paper]{article}
\usepackage[dvipsnames]{xcolor} %added
\usepackage[hyperref]{acl2020}
\usepackage{times}
\usepackage{latexsym}

\usepackage{amsmath}
\usepackage{amsfonts}
\usepackage{amssymb}
\usepackage{graphicx}  
\usepackage{multirow}
\usepackage{booktabs}
\usepackage{pifont}
\usepackage{comment}
\usepackage{caption,subcaption} %?? allowed
\newcommand{\E}{\mathbb{E}}
\let\oldding\ding% Store old \ding in \oldding
\renewcommand{\ding}[2][1]{\scalebox{#1}{\oldding{#2}}}% Scale \oldding via optional argument

\usepackage{microtype}

\aclfinalcopy % Uncomment this line for the final submission

\newcommand{\eat}[1]{}
\newcommand*{\affaddr}[1]{#1} % No op here. Customize it for different styles.
\newcommand*{\affmark}[1][*]{\textsuperscript{#1}}
\newcommand*{\email}[1]{\texttt{#1}}

\title{Generating Narrative Text in a Switching Dynamical System}

\author{%
Noah Weber \affmark[3,1], Leena Shekhar \affmark[1]\thanks{\hspace{0.1cm}Author now at Microsoft}\hspace{0.10cm}, Heeyoung Kwon\affmark[1] \\ 
\textbf{Niranjan Balasubramanian}\affmark[1], \textbf{Nathanael Chambers} \affmark[2] \\
\affaddr{\affmark[1]Stony Brook University \hspace{0.3cm}} \\
\email{\{lshekhar,heekwon,niranjan\}@cs.stonybrook.edu}\\
\affaddr{\affmark[2]United States Naval Academy \hspace{0.3cm}}\\
\email{nchamber@usna.edu} \\
\affaddr{\affmark[3]Johns Hopkins University \hspace{0.3cm}}\\
\email{nweber6@jhu.edu} 
}

\date{}
\begin{document}
\maketitle
\begin{abstract}
Early work on narrative modeling used explicit plans and goals to generate stories, but the language generation itself was restricted and inflexible. Modern methods use language models for more robust generation, but often lack an explicit representation of the scaffolding and dynamics that guide a coherent narrative.
This paper introduces a new model that integrates explicit narrative structure with neural language models, formalizing narrative modeling as a \emph{Switching Linear Dynamical System} (SLDS). A SLDS is a dynamical system in which the latent dynamics of the system (i.e. how the state vector transforms over time) is controlled by top-level discrete \emph{switching} variables. The switching variables represent narrative structure (e.g., sentiment or discourse states), while the latent state vector encodes information on the current state of the narrative.
This probabilistic formulation allows us to control generation, and can be learned in a semi-supervised fashion using both labeled and unlabeled data. Additionally, we derive a Gibbs sampler for our model that can ``fill in" arbitrary parts of the narrative, guided by the switching variables.
Our filled-in (English language) narratives outperform several baselines on both automatic and human evaluations.
\end{abstract}

%auto-ignore
\section{Introduction}
A narrative is a textualized sequence of events that serves as a coherent outline for an actual story~\cite{prince2003dictionary}. Effective narratives are typically built on top of higher level \emph{narrative scaffolds}\footnote{We use the term scaffold as an umbrella term to cover many types of plans and structures that underlie stories.} which specify at an abstract level how the story should evolve along different dimensions. Examples of these scaffolds include descriptions of the emotional trajectory of a story \cite{vonnegut, reagan2016emotional}, the goals of characters throughout the story \cite{Meehan77tale-spin, Turner93}, or the abstract types of events that may occur \cite{DBLP:journals/corr/MartinAHSHR17}. The parts of a scaffold are generic, and can be reused across stories. 

To be better utilized in creating or describing a particular instance of a story, each element in the scaffold should be associated with a description of \textit{how} it changes some property of the particular story state (e.g, how to transition a story state from positive to negative sentiment). We refer to these transitions as \emph{narrative dynamics}. 

\begin{table}[t!]
\centering
\begin{tabular}{c}
\toprule
\textcolor{gray}{\small Tom didn't know why his internet speed was so slow.} \\
\textbf{\small Tom wasn't sure what to do with his computer.} \\
\textbf{\small He thought he would fix it himself.} \\
\textbf{\small Tom was surprisingly good.} \\
\textcolor{gray}{\small Tom was happy to be surfing the internet again} \\ 
\bottomrule
\end{tabular}
\caption{A sample filled in narrative generated by our SLDS model given the first and last sentences as input (grayed out), the middle 3 sentences are imputed by our model (bold).}
\label{tab:init_story}
\end{table}

Prior work on automatic narrative generation has a rich history of modeling both \emph{narrative scaffolds} and \emph{narrative dynamics} \cite{Meehan77tale-spin,lebowitz1985story,Turner93,Riedl2006,Riedl2010}.
The modeling of both narrative scaffold and dynamics often imbued these systems with a greater degree of control for the user in generating stories, allowing users to flexibly specify desired outcomes or plot points (or more generally, the state of the narrative) that should be achieved at certain sections of the story. 
Though successful in this regard, this success has only been realized in closed domains, where the narrative scaffolds can be specified in a limited ontology and the dynamics operations can be written by hand (such as e.g. the action schemata of \citet{Riedl2010}). Neural generation has since helped scale to open domains \cite{roemmele2015creative,khalifa2017deeptingle} but not with the same level of control over the narrative. Several recent works have looked at adding the narrative scaffolding component back into neural text generating systems \cite{Fan2018,DBLP:journals/corr/MartinAHSHR17,planandwrite,skeleton}. These systems however still do not utilize an explicit model of narrative dynamics, and are thus restricted in the controllability aspect.

Building controllable methods with modern language generation requires mechanisms that can meet the targets set by the narrative scaffold as well as the constraints on the narrative dynamics. 
For instance, suppose we want to generate stories constrained to meet specified goals, such as being constrained to start and end with specified sentences (such as those given in the narrative in Table~\ref{tab:init_story}). 
While conditioning on a scaffold alone can help generate intervening sentences, it does not guarantee that these intervening sentences transition in a natural way that cohesively leads into the last sentence. Modeling the dynamics of a narrative separately may allow the system to infer an optimal path through the narrative in order to reach the specified goal. This is analogous to how plan-based generation systems used search algorithms over narrative actions to generate a feasible path to a goal \cite{Riedl2010}.

In this work, we show how the insight of modeling the structure of a narrative along with general purpose dynamics can be combined with modern neural network based language models. We do this by explicitly modeling the narrative state with a latent vector, and modeling how this state transforms over time as a \textit{Switching Linear Dynamical System} (SLDS). We show how this formulation neatly captures the concepts of narrative dynamics and scaffolds in a way compatible with current neural generation systems, and demonstrate how the model can flexibly be used both for standard language modeling and controlled generation of narratives. Under our model, the task of ``filling in" a narrative conditioned on arbitrarily specified sections can be recast as sampling from a specific class of conditional probability distributions. We show how this distribution can be sampled from via Gibbs sampling, permitting our model to flexibly ``fill in" arbitrary parts of a narrative without being trained to do so. We evaluate our model with both human evaluation and several automatic measures\footnote{Evaluation in this paper is done on English text data} and show that our model outperforms several strong baselines. Model code is available. \footnote{\texttt{github.com/StonyBrookNLP/SLDS-Stories}} 

%\nb{How about enumerating some contributions? Given that the AAAI reviewers can be a diverse pool, it might be best to spell out the contributions and their impact potential.}

%We show how this formulation allows our model to flexibly ``fill in" a narrative conditioned on arbitrarily specified sections.

%\nb{Say why gibbs is needed. To marks point also say what it takes to generate from this model --- that it requires sampling and that it needs to be done in a particular way etc.} Furthermore, we derive an (approximate) Gibbs sampling algorithm which allows our model to flexibly ``fill in" a narrative conditioned on arbitrarily specified sections.

\section{A Switching Dynamical System for Narrative Generation}
In this section, we give a brief overview of Switching Dynamical systems and how they can be used to capture both a scaffold of the narrative as well as the narrative dynamics. We then describe in detail the components of our model and its relation to existing models. 
\subsection{Narrative Dynamics in a Dynamical System} \label{subsec:lds}
The specifics of the narrative (characters, setting, etc.), will differ between stories, but as \citet{propp2010morphology} notes, the way they transition to the next point in the narrative (what we refer to as ``narrative dynamics") is often shared. Let's say that, as done often, we represent the `narrative specifics' at time step\footnote{In our case, we take each sentence in the narrative to be a different timestep. Different levels of granularity for a timestep may be more befitting for other domains.} $i$ with a latent vector $Z_i$. A natural way to explicitly model how this state evolves over time that fits with the above observation is as a \textit{Linear Dynamical System}:
\begin{align*}
Z_{i +1} = AZ_{i} + \epsilon \text{ ; } \epsilon \sim \mathcal{N}(0, \Sigma)
\end{align*}
Where $A$ is a matrix, shared across all narratives, and $\Sigma$ is a noise term that takes into consideration idiosyncrasies different narratives will have\footnote{Note that a bias term may also be added here (we do this in our implementation). We leave the bias off here for clarity}. The fact that the shared transition matrix $A$ is linear means that narratives will have linearly analogous trajectories through time, despite having different details (comparable to stories with different settings but matching structures such as \textit{Ran}/\textit{King Lear}, \textit{Ulysses}/\textit{Odyssey}, etc). Of course, the fatal flaw of the model is that it assumes there exists only one transition matrix, and thus only one possible way to transition through a narrative!

\subsection{Narrative Scaffolds as Switching Variables}
A more fitting model would thus be a \textit{Switching Linear Dynamical System} \cite{Ackerson1970,Chang1978,murphy1998switching}. In an SLDS, we assume there exists a set of $K$ different sets of dynamics, $\{(A_1, \Sigma_1),...(A_K,\Sigma_K)\}$. At time step $i+1$, \textit{one} of these sets of dynamics is used. The one used depends on the value of a discrete variable at time step $i+1$ called the switching variable, $S_{i+1} \in \{1,...K\}$:
\begin{align*}
Z_{i+1} = A_{S_{i+1}}Z_{i} + \epsilon \text{ ; } \epsilon \sim \mathcal{N}(0, \Sigma_{S_{i+1}})
\end{align*}
There is a switching variable $S_i$ associated with each time step. The switching variable value itself evolves over time by a prior Markov process, $P(S_{i+1} | S_{i})$\footnote{Other ways to formulate this transformation are also possible \cite{barber2006expectation,linderman2016recurrent}. The Markov assumption is a common one and we use it here for simplicity.}. This top level chain of switching variables thus forms our \textit{narrative scaffold}, indicating \textit{what} transitions we must go through in the narrative, with the dynamics matrices indicating \textit{how} they transition. 

\subsection{Narrative Scaffold - Emotional Trajectory}
What the switching variables actually represent can be chosen by the user. Straightforward narrative scaffolds include event sequences \cite{DBLP:journals/corr/MartinAHSHR17}, keywords \cite{planandwrite}, or latent template ids \cite{wiseman2018learning}. More complex but potentially more informative scaffolds may be created using concepts such as story grammar non-terminals \cite{lakoff1972structural,thorndyke1977cognitive}, or  character action taken throughout a story \cite{Riedl2010NarrativePB}.

In our work, we use the sentiment trajectory of the narrative as the scaffold. That is, each $S_i$ for a sentence indicates the overall coarse sentiment of the sentence (Positive, Negative, or Neutral). Though simple, the overall sentiment trajectory of a narrative is important in defining the high level `shape' of a narrative often shared among different narratives \cite{vonnegut,reagan2016emotional}. Furthermore, sentiment trajectory has been shown to be fairly useful in story understanding tasks \cite{chaturvedi2017story,liu2018narrative}. We discuss in the conclusion future directions for using different types of scaffolds.

\subsection{The Full Model}
The final component of the model is a conditional language model that generates sentence $i$ conditioned on the current $Z_i$, and all previous sentences, $X_{:i}$. Generation continues until an \texttt{<eos>} is reached. This conditional language model may be parameterized as desired, but in this work, we parameterize it as an RNN neural network language model.

\begin{figure}[h!]
\centering
\includegraphics[scale=0.5]{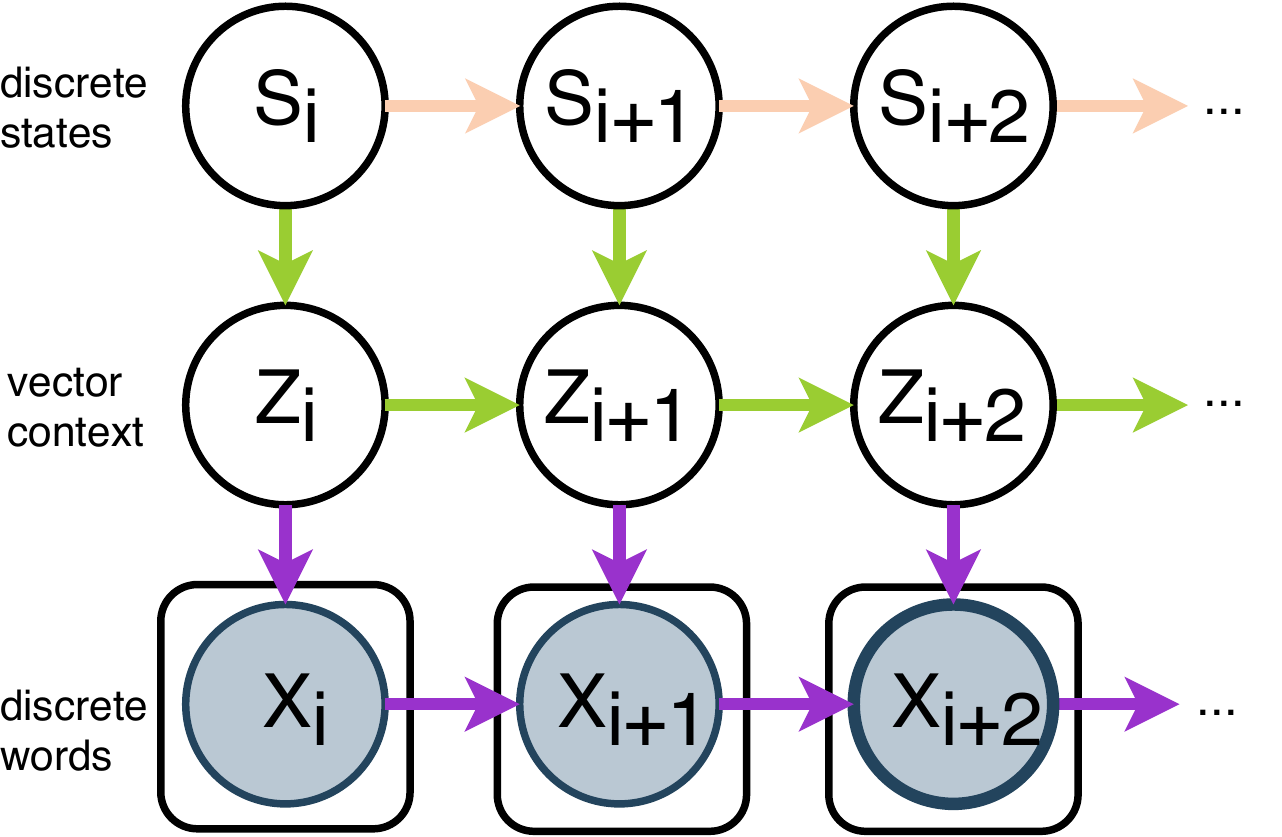}
\caption{SLDS Generative model-$S_i$ is a discrete state (sentiment of a sentence in a multi-sentence narrative ). $Z_i$ is a continuous latent vector that is conditioned on to generate the $i$th sentence in the narrative , $X_i$. The dynamics of the  narrative are completely captured in the dynamical system controlling the latent vector $Z$. How to transition from $Z_i$ to $Z_{i+1}$ is determined by the state variable $S_{i+1}$. Arrows from $X_i$ to $X_{i+2}$ have been left out for clarity.}
\label{fig:slds}
\end{figure}

The graphical model for our SLDS is pictured in Figure~\ref{fig:slds}. The model consists of three sets of variables: (1) Switching variables $S_1,...,S_N$, (2) Latent state variables $Z_1,...,Z_N$ capturing the details of the narrative at sentence $i$, (3) The sentences themselves $X_1,...X_N$, where each sentence $X_i$ has $n_i$ words, $x^i_1,...x^i_{n_i}$. The joint over all variables factorizes as below into the following components ($X_{:i}$ stands for all sentence before $X_i$):

\begin{align*}
P(\textbf{S}, \textbf{Z}, \textbf{X}) = (\prod^N_i \underbrace{P(X_i | Z_i, X_{:i})}_{{\color{DarkOrchid}\text{\ding[1.3]{184}}}}) \\ (\prod^N_i \underbrace{P(Z_i | Z_{i-1}, S_i)}_{{\color{YellowGreen}\text{\ding[1.3]{183}}}}) (\prod^N_i \underbrace{P(S_i | S_{i-1})}_{{\color{Apricot}\text{\ding[1.3]{182}}}})
\end{align*}
{\color{Apricot}\text{\ding{182}}} \textbf{Narrative Scaffold Planner}: The factor $P(S_i | S_{i-1})$ is a transition matrix, which we calculate via count based statistics from training. It is fed in as prior knowledge and fixed.

\noindent{\color{YellowGreen}\text{\ding{183}}} \textbf{Narrative Dynamics Network}: The factor $P(Z_i | Z_{i-1}, S_i)$ is determined like a switching linear dynamical system:
\begin{align*}
Z_i = A_{S_i}Z_{i-1} + B_{S_i}\epsilon, \text{ ; } \epsilon \sim \mathcal{N}(0, I)
\end{align*}
which is equivalent to drawing $Z_i$ from a Normal distribution with mean $A_{S_i}Z_{i-1}$ and variance $B_{S_i}B_{S_i}^T$.

\noindent{\color{DarkOrchid}\text{\ding{184}}} \textbf{Conditional Language model}: The factor $P(X_i | Z_i, X_{:i})$ is parameterized by an RNN language model conditioned on the latent $Z_i$. 

\section{Learning and Posterior Inference}
% We learn the model parameters by maximizing a lower bound on the likelihood:
%\[
%\log{p(X)} \geq \E_{q(S,Z | X)}\Big[\frac{\log{p(X, S, Z)}}{\log{q(S,Z|X)}}\Big]
%\]

%\begin{figure}[h!]
%\centering
%\includegraphics[scale=0.5]{figures/inferslds_thick.pdf}
%\caption{Inference network graphical model. The dependencies between past and future sentences and each $Z_i$ and $S_i$ have been left out for clarity}
%\label{fig:inferslds}
%\end{figure}

Due to the conditionals parameterized by neural networks we use amortized variational inference in a manner similar to Variational AutoEncoders \cite{kingma2013auto}, both to learn an approximate posterior $q(S, Z | X)$ and to learn the generative model parameters by maximizing a lower bound on the 
data likelihood (ELBO). We assume that the approximate posterior factorizes as follows:
\begin{align*}
q(\textbf{S}, \textbf{Z} |&\textbf{X}) = \\
&(\prod^N_i q(S_i | \textbf{X})) (\prod^N_i q(Z_i| Z_{i-1}, S_i, X_{:i}, X_{i}))
\end{align*}
Like in VAEs, computing these individual factors is done through a parameterized function called the \textit{inference} or \textit{recognition} network whose parameters are trained jointly with the generative model. In our case there are 
two forms for the factors in our posterior: (1) The first form, $q(S_i | \textbf{X}) = q_{S_i}$ is parameterized by a classifier that takes in the set of sentences $\mathbf{X}$ and outputs a categorical distribution over the switching variables. (2) The second form, $q(Z_i| Z_{i-1}, S_i, X_{:i}, X_{i}) = q_{Z_i}$ is realized by functions $f_{\mu}(Z_{i-1}, S_i, X_{:i}, X_{i})$ and $f_\sigma(Z_{i-1}, S_i, X_{:i}, X_{i})$ that output the mean and variance, respectively, of a Gaussian over $Z_i$. 

Borrowing terminology from VAEs, the approximate posterior (the factors given above) act as an `encoder', while the generative model from the previous section can be seen as the `decoder'. This type of training has been previously used in \cite{Krishnan2015,Krishnan2017,Fraccaro2016,Fraccaro2017,Karl2016}. %these types of models

\subsection{Lower bound formula \& exact training algorithm}
As mentioned previously, we optimize all parameters (including the variational factor functions) by optimizing a lower bound on the data likelihood. The model may be trained either with supervision labels for the switching states (in our case, sentiment labels) or without supervised labels. 

If one is training \textit{without the sentiment labels}, then the lower bound on the marginal likelihood (and thus our optimization objective) may be written as follows:

\begin{equation*}
L = \E_{S_1,..S_N \sim q_{S_i}} \Bigg[ M - \sum_i^N KL(q_{S_i} || p(S_i | S_{i-1})) \Bigg]
\end{equation*}

\begin{align*}
\text{where, } M &= \E_{Z_1,..Z_N \sim q_{Z_i}} \Big[\sum_i^N (\log{p(X_i | Z_i)} \\
  & - KL(q_{Z_i} || p(Z_i | Z_{i-1}, S_i))) \Big]
\end{align*}
The derivation for this objective is identical to that found in \cite{Krishnan2017,Fraccaro2016}, and simply relies on using properties of iterated expectations. All expectations are estimated with Monte Carlo samples.

If training \textit{with the sentiment labels} $S_1,...,S_N$, then the objective is similar (but without the sampling of the switching states), and is augmented with an additional supervision objective as done in \citet{kingma2014semi}: %(one per sentence)

\begin{align*}
L_S &= M + \sum_i^N q_{S_i}
\end{align*}

Final training procedure for a single  narrative is:
\begin{enumerate}
    \item For each sentence (starting from the first), sample the switching state  $S_i$ from $q(S_i | \textbf{X})$.
    \item For each sentence (starting from the first), sample the latent $Z_i$ from $q(Z_i | S_i, Z_{i-1}, X)$.
    \item Evaluate the data likelihood and KL term(s) with these samples.
    \item Take the gradients of the objective function w.r.t. all parameters, using the reparameterization trick for $q_{Z_i}$ \cite{kingma2013auto} or the Gumbel-Softmax\footnote{We use the soft variant of Gumbel-Softmax. Rather than forcing a hard choice for $S_i$, we directly use the Gumbel-Softmax output and combine the transition matrices via a convex combination} trick for $q_{S_i}$ \cite{Jang2017}, and optimize.
\end{enumerate}

\section{Interpolations via Gibbs Sampling} \label{sec:gibbs}
One of the benefits of probabilistic formulation is the possibility (if an inference procedure can be found) of generating narratives with specific constraints, where the constraints may be specified as clamped variables in the model. In this section, we show how narratives may be generated conditioned on arbitrary bits and pieces of the narrative already filled in, using approximate Gibbs sampling. This allows one to, for example, interpolate a  narrative given the first and the last sentence (similar to how earlier story generation systems were able to generate with a given end goal in mind). Some examples of these interpolations generated by our system can be found in Table~\ref{tab:sample-stories}. We give the equations and summarize the algorithm in the next sections. 

\subsection{Conditionals for Gibbs Sampling}
For our Gibbs sampling algorithm we give the narrative scaffold (switching variables), $S_1,...,S_T \in \mathbf{S}$ and a set of observed sentences, $\mathbf{X^+}$. This may be any set of sentences (the first and last, just the second sentence, etc) as inputs to the system. We wish to find values for the unobserved sentences in set $\mathbf{X^-}$ by sampling from the distribution $P(\mathbf{X^-}, Z_1,...,Z_T | \mathbf{S},\mathbf{X^+})$. We perform this sampling via Gibbs sampling. Two different forms of conditionals need to be derived to do Gibbs sampling. One over some $Z_i$ conditioned on everything else, and one over some $X_i$ conditioned on everything else. 

By using the d-separation properties of the graph, and substituting the true posterior over $Z_{i}$ with our approximate posterior $q$, we can show the first distribution is approximately proportional to \footnote{See the Supplemental Material for a full detailed derivation of the conditionals in this section}  
\begin{align*}
&P(Z_i | Z_{i-1}, Z_{i+1}, S_{i}, S_{i+1}, X_i, X_{i-1})  \\
&\propto P(Z_{i+1} | S_{i+1}, Z_{i}) P(Z_{i} | Z_{i-1}, S_{i}, X_{i}, X_{i-1}) \\
&\approx P(Z_{i+1} | S_{i+1}, Z_{i}) q(Z_{i} | Z_{i-1}, S_{i}, X_{i}, X_{i-1}) 
%& = \mathcal{N}_{Z_{i+1}}(A_{S_{i+1}}Z_i,\Sigma_{S_{i+1}}) \mathcal{N}_{Z_{i}}(f_{\mu}(\cdot),f_{\sigma}(\cdot))
\end{align*}
The last line is the product between a Gaussian density over $Z_{i+1}$ and $Z_{i}$, respectively. With some algebraic manipulations, one can show the last line is proportional to a single Gaussian PDF over $Z_i$:

\begin{align}
\mathcal{N}_{Z_{i+1}}&(A_{S_{i+1}}Z_i,\Sigma_{S_{i+1}}) \mathcal{N}_{Z_{i}}(f_{\mu}(\cdot),f_{\sigma}(\cdot)) \\
&\propto \mathcal{N}_{Z_{i}}(\mu_*,\Sigma_*) \hphantom{1}\text{where,} \nonumber \\
\Sigma_* = &\big(A_{S_{i+1}}^T \Sigma_{S_{i+1}}^{-1} A_{S_{i+1}} + f_{\sigma}(\cdot)^{-1}\big)^{-1}\nonumber   \\
\mu_* = &\Sigma_*^T\big(Z_{i+1} \Sigma_{S_{i+1}}^{-1} A_{S_{i+1}} + f_{\mu}(\cdot)^T f_{\sigma}(\cdot)^{-1}\big)^{T}\nonumber  
\end{align}

To find the second conditional, one can use the d-separation properties of the graph to find that it is proportional to: % (where $X_{:i}$ stands for all sentence before $X_i$):
\begin{align*}
&P(X_i | Z_{i}, Z_{i+1}, S_{i}, S_{i+1}, X_{:i}, X_{i+1})  \\
&\propto P(X_{i+1} | X_{:i}, X_i, Z_{i+1}) P(X_{i} | X_{:i}, Z_{i}) 
\end{align*}
These two distributions are simply factors of our conditional language model, and both terms can thus be evaluated easily. In theory, one could use this fact to sample the original conditional via Metropolis-Hastings \footnote{MH sampling has also been used for story generation in \citet{harrison2017toward}, albeit in a different fashion}. Unfortunately, we found this approach to be much too slow for practical purposes. We observed that the simple heuristic of deterministically assigning $X_i$ to be the greedy decoded output of the conditional language model $P(X_{i} | X_{:i}, Z_{i})$ works well, as evidenced by the empirical results. We leave it for future work to research different conditional language model parameterizations that allow easy sampling from this conditional\footnote{One possibility is to use `orderless' pretrained Transformer models such as BERT\cite{Devlin2018BERTPO}, in which it has been shown possible to sample from the Gibbs conditional \cite{Wang2019}}

\subsection{Gibbs Sampling Interpolation Overview}
The variables in the Gibbs sampler are first initialized using some heuristics (see Supplemental Materials for details). After initialization, performing the interpolations with Gibbs sampling follows the below two step process:
\begin{itemize}
\item For each $Z_i$, sample a value $Z^\prime$ from equation $(1)$ and set $Z_i$ to $Z^\prime$.
\item For each $X_i$ in $\mathbf{X}^-$, find a new value for $X_i$ by running greedy decoding using the conditional language model.
\end{itemize}

\section{Training Details}

\subsection{Dataset and Preprocessing}
We use the ROCStories corpora introduced in \citet{MostafazadehCHP16}. It contains 98,159 short commonsense stories in English as training, and 1,570 stories for validation and test each. 
Each story in the dataset has five-sentences and captures causal and temporal commonsense relations. We limit our vocabulary size to 16,983 based on a per-word frequency cutoff set to 5. 
For sentiment tags, we automatically tag the entirety of the corpus with the rule based sentiment tagger, Vader \cite{hutto2014vader}, and bucket the polarity scores of Vader into three tags: neutral, negative, and positive. These tags form the label set of the $S$ variables in our SLDS model. We tokenize the stories with Spacy tokenizer\footnote{spacy 2.1.4}. Each sentences in the input  narrative has an \texttt{<eos>} tag except for the S2S model discussed below.

\subsection{Switching Linear Dynamical System (SLDS)}
SLDS has RNN encoder and decoder networks with single layer GRU cells of hidden size 1024. Model uses an embedding size of 300. We train the model using Adam optimizer with the defaults used by PyTorch. We stop training the models when the validation loss does not decrease for 3 consecutive epochs. Training details remain same as above unless otherwise mentioned. 

\subsection{Baselines}

\begin{itemize}
\item Language Model (LM): We train a two layer recurrent neural language model with GRU cells of hidden size 512.

\item Sequence-to-Sequence Attention Model (S2S): We train a two layer neural sequence to sequence model equipped with bi-linear attention function with GRU cells of hidden size 512. Sentiments tags for a narrative (1 for each sentence) are given as input to the model and the corresponding sentences are concatenated together as the output with only one \texttt{<eos>} tag at the end. This model is trained with a 0.1 dropout. This model is comparable to the static model of \cite{planandwrite}, and other recent works employing a notion of scaffolding into neural generation (albeit adapted for our setting).

\item Linear Dynamical System (LDS): We also train a linear dynamical system as discussed in Section~\ref{subsec:lds} as one of our baselines for fair comparisons. Apart from having just a \textit{single} transition matrix this model has the same architectural details as SLDS. 

\item Semi-Supervised SLDS (SLDS-X\%): To gauge the usability of semi-supervision, we also train semi-supervised SLDS models with varying amount of labelled sentiment tags unlike the original model which uses 100\% tagged data. We refer to these as SLDS-X\%, where \textit{X} is the \% labelled data used for training: 1\%, 10\%, 25\%, and 50\%. 
\end{itemize}

\section{Evaluations}
As described above, our model is able to perform  narrative interpolations via an approximate Gibbs sampling procedure.  
At the core of our evaluations is thus a fill-in-the-sentences task.
We provide 1 or 2 sentences, and require the model to generate the rest of the narrative .
We evaluate this via automatic evaluations as well as with crowd-sourced human evaluations. We also report perplexity to evaluate the models' ability to fit the data. Lastly, we look at whether the transitions learned by the SLDS models capture what they are intended to capture: does using the transition matrix associated with a sentiment tag (positive/negative/neutral) lead to a generated sentence with that sentiment?

\subsection{Generating the Interpolations} \label{sec:interp}
For the SLDS models, the interpolations are generated via the Gibbs sampling algorithm described earlier. In all experiments for the SLDS models we draw \textbf{50} samples (including burn in samples) and output the interpolation that maximizes the probability of the given sentence(s).
Since the baselines do not have the means for doing interpolations, we simulate `interpolations' for the baselines; we draw \textbf{1000} samples using top k (with k=15) truncated sampling (conditioned on the given initial sentences, if available). We then output the sample that maximizes the probability of the clamped sentences around which we are interpolating the others. We allow the S2S access to the gold sentiment tags. To give a lower bound on the performance of the SLDS model, we do not provide it with gold tags. We instead provide the SLDS model with the semi-noisy tags that are output from $q(S_i | X)$.\footnote{Note that missing sentences, $X$, are used \textit{only} for computing these noisy tags}

\begin{table*}[h!]
\centering
\begin{tabular}{l|l|r|r|r|r|r|r|r|r}
\toprule
\# Sent(s) & Metric & SLDS & SLDS-1 & SLDS-10 & SLDS-25 & SLDS-50 & S2S & LM & LDS \\ \toprule
\multirow{ 4}{*}{$2^{nd}$} & R1 & 17.60 & 19.36 & 20.46 & \textbf{20.92} & 19.55 & 18.79 & 18.30 & 17.33 \\
& R2 & 2.43 & 3.46 & 3.86 & \textbf{4.10} & 3.43 & 3.14 & 2.83 & 2.31 \\
& RL & 16.43 & 17.76 & 19.03 & \textbf{19.45} & 17.87 & 17.39 & 16.68 & 15.97 \\ 
& M & 6.35 & 6.84 & 6.98 & \textbf{7.15} & 6.94 & 7.11 & 6.76 & 6.13 \\ 
\midrule
\midrule
\multirow{ 4}{*}{$4^{th}$} & R1 & 16.98 & 17.90 & 18.64 & 18.14 & \textbf{19.39} & 15.38 & 14.03 & 17.06 \\
& R2 & 2.20 & 2.61 & 2.74 & 2.29 & \textbf{3.23} & 1.97 & 1.40 & 2.31 \\
& RL & 15.20 & 16.21 & 16.69 & 16.08 & \textbf{17.43} & 13.90 & 12.64 & 15.24 \\ 
& M & 6.33 & 7.11 & 6.92 & 6.53 & \textbf{7.18} & 5.84 & 5.61 & 6.89 \\ 
\midrule
\midrule
\multirow{ 4}{*}{$1^{st}+2^{nd}$} & R1 & 15.40 & 16.04 & 16.11 & 16.33 & 16.27 & \textbf{18.91} & 17.38 & 14.32 \\
& R2 & 1.79 & 1.65 & 1.97 & 2.17 & 1.83 & \textbf{2.62} & 2.03 & 1.47 \\
& RL & 14.63 & 15.15 & 15.23 & 15.47 & 15.27 & \textbf{17.89} & 16.48 & 13.41 \\ 
& M & 5.34 & 5.27 & 5.40 & 5.44 & 5.42 & \textbf{6.81} & 6.07 & 4.80 \\ 
\midrule
\midrule
\multirow{ 4}{*}{$3^{rd}+4^{th}$} & R1 & 23.35 & 23.59 & 23.57 & \textbf{23.65} & 23.60 & 20.68 & 20.01 & 21.66 \\
& R2 & 3.77 & 3.35 & 3.76 & 3.58 & \textbf{3.93} & 2.51 & 1.91 & 2.94 \\ 
& RL & 21.56 & 21.49 & 21.87 & \textbf{21.88} & 21.67 & 18.87 & 18.28 & 20.04 \\ 
& M & 8.28 & 8.26 & 8.22 & 8.12 & \textbf{8.29} & 7.51 & 7.26 & 7.87 \\ 
\bottomrule
\end{tabular}
\caption{F1 scores for ROUGE-1, 2, and L and METEOR (M) (default mode score) for randomly sampled 500 stories from the test set. \#Sents(s) column represents the ``fill in" sentence(s) that the models generated using Gibbs sampling. Our SLDS models pick the best of \textbf{50} samples, the baselines models pick the best of \textbf{1000} samples}
\label{tab:auto-interp}
\end{table*}

\subsection{Automatic Evaluation of Interpolations}
We automatically evaluate on four different types of interpolations (where different combinations of sentences are removed and the model is forced to regenerate them),
We evaluate the generations with the ROUGE \cite{Lin2004ROUGEAP}\footnote{Pyrouge package at \url{pypi.python.org/pypi/pyrouge/0.1.3}.} and METEOR \cite{meteor} metrics\footnote{E2E NLG Challenge scoring scripts at
\url{https://github.com/tuetschek/e2e-metrics}.
} using the true sentences as targets. Table~\ref{tab:auto-interp} shows the automatic evaluation results from interpolations using our proposed models and baselines. The \#Sent(s) column indicates which sentence(s) were removed, and then regenerated by the model. We gave the baselines a slight edge over SLDS because they pick the best out of 1000 samples while SLDS is only out of 50. The SLDS models see their largest gain over the baseline models when at least the first sentence is given as an input. The baseline models do better when the first and second sentence need to be imputed. This is likely due to the fact that having access to the earlier sentences allows a better initialization for the Gibbs sampler. Surprisingly, the semi-supervised variants of the SLDS models achieve higher scores. The reasons for this is discussed below in the Perplexity section.

\begin{table*}[h!]
\small
\centering
\begin{tabular}{l|l}
\toprule
\textcolor{gray}{Yesterday I was at the mall shopping.} & \textbf{I had a bad relationship with my girlfriend.} \\
\textbf{Suddenly, a man in a mask appeared.} & \textbf{She was very upset and told me she was pregnant.} \\
\textbf{The man was walking around and he was gone.} & \textcolor{gray}{Rachel told her family and it was very difficult.}\\
\textbf{He had to call me to help me get out of the store.} & \textcolor{gray}{They all cried.}\\
\textcolor{gray}{I heard on the news that night they sent the UNK to prison.} & \textcolor{gray}{They agreed that on the whole, it was good news.} \\
\midrule
\textcolor{gray}{Ed was playing baseball in his yard.} & \textbf{Last week I had an idea.} \\
\textbf{He was running down the hill.} & \textbf{I was so nervous that I decided to make a presentation.} \\
\textbf{His ball was coming towards him.} & \textcolor{gray}{I soon found it hard to come up with new ideas.}\\
\textbf{It was very scary!} & \textcolor{gray}{I didn't think it would be so hard.}\\
\textcolor{gray}{Ed was scared.} & \textcolor{gray}{But then, an idea came to me and I was back on track.} \\
\midrule
\textcolor{gray}{Ben has always wanted to learn how to play the piano} & \textcolor{gray}{Tim was always on his bike during the summer.} \\
\textbf{His parent bought him one.} & \textbf{He had a lot of fun.} \\
\textcolor{gray}{Ben enrolls in a piano class with a local tutor.} & \textbf{One day he decided to cut his bike down.} \\
\textbf{Ben practiced every day.} & \textbf{He hit a rock and fell off the bike and hit a tree.} \\
\textcolor{gray}{He gets better with every lesson.} & \textcolor{gray}{He broke his arm.} \\
\bottomrule

\end{tabular}
\caption{Sample interpolations from Gibbs sampling. Grayed out lines are provided as input and bold sentences are generated by SLDS.}
\label{tab:sample-stories}
\end{table*}

\subsection{Human Evaluation of Interpolations} \label{semi_super}
\subsubsection{Annotation Scheme}
As automatic evaluation metrics are not sufficient to assess the quality of any creative task such as  narrative generation, we measure the quality of the generations through human evaluation of 200 stories on the Amazon Mechanical Turk platform. 
We provided Turkers with two generated narratives from two different models, each with five sentences. The first and last sentences were fed to each model as input, and the middle three sentences were generated. Each pair of narratives is graded by 3 users each with two tasks: (1) to rank on a scale of 0-3 each of the sentences except the first one on the basis of its coherency with the previous sentence(s) and (2) compare and rank the two narratives based on their overall coherency, ie how well the story connects the starting/ending sentences.

\subsubsection{Human Evaluation Results}
Table~\ref{tab:human-interpl} reports the result of human evaluations of SLDS and baseline generations. We can observe that people preferred narratives generated by SLDS over the ones generated by baseline models (LM and S2S) as they found the former model more coherent, which is an important criteria for narrative generation. 
%SLDS was reported better \textbf{51.3\%} of the time, 35.0\% for the LM, and 13.7\% as a tie. The mean sentence level coherence score for SLDS is not only higher than the baseline, but also with a smaller standard deviation, which supports our coherence hypothesis.
\textbf{51.3\%} of the time SLDS generates better narratives than the LM model while LM in turn does it only \textbf{35.0\%} of the times. 13.7\% of the generations end up in tie. The mean sentence level coherence score for SLDS is around 12.5\% larger than that of the LM, with a slightly lower standard deviation. We see similar results when compared against the S2S model.

\begin{table}[ht]
\centering
\begin{tabular}{l|r|r}
 System & Sent Coh.\ (0-3) & Best Story\\
\toprule
LM & 1.68 \small{$\pm$ 1.01} & 35.0\% \\  
SLDS & \textbf{1.89 \small{$\pm$ 0.96}} & \textbf{51.3\%} \\  
\midrule
S2S & 1.67 \small{$\pm$ 1.00} & 35.1\% \\  
SLDS & \textbf{1.87 \small{$\pm$ 0.97}} & \textbf{51.9\%} \\  
\end{tabular}
\caption{Human evaluation scores for filled-in  narrative generation. Humans judged sentence coherence and chose which model filled in the \textit{most} coherent  narrative overall (13.7\% and 13\% tie for LM and S2S).}
\label{tab:human-interpl}
\end{table}

\subsection{Language Modeling Perplexity Score} \label{subsec:ppl}
As our models are essentially language models, we evaluated their per-sentence negative log-likelihood and per-word perplexity scores\footnote{Note that since S2S appends the \texttt{eos} token only at the end, its per-sentence NLL is slightly lower than that of LM.}, which can be viewed as an indirect measure of how well a system works as a generative model of narrative text. For the SLDS and LDS models these scores are approximations, an upper bound (the negative of the ELBO) to the actual values. For the other two models the scores are exact. A good model should assign low perplexity scores to its test set. 
In Table~\ref{tab:perplex} SLDS achieves the lowest scores, implying that it is able to model the data distribution well. 
In Table~\ref{tab:eval-semis} we also calculate the perplexity scores for the semi-supervised SLDS models to assess the effectiveness of semi-supervised training. Surprisingly, the models with less supervision scored better in terms of perplexity. One possibility for this might be the use of the soft Gumbel-Softmax in the semi-supervised models. The soft Gumbel-Softmax variant does not commit to using a single transition matrix at each time step (instead linearly combining them, weighted by the Softmax weights). This fact may permit the model greater flexibility in fitting the training data. While this leads to better scores in metrics such as perplexity or BLEU, it does leads to transitions that are worse in capturing the properties they should be capturing, as we shall see in the next section. %These scores are calculated as discussed in the previous section. %S2S model does not have a \texttt{<eos>} tag at the end of each sentence in a story unlike the other models thus has lesser number of total words. 

\begin{table}[h]
\centering
\begin{tabular}{l|r|r} 
System & NLL & PPL \\
\toprule
 LM &  196.30 & 35.41 \\ 
 S2S &  192.25 & 43.36 \\
 LDS & $\leq$186.24 & 29.49 \\ 
 SLDS & $\leq$\textbf{182.17} & \textbf{27.39} \\
\end{tabular}
\caption{NLL and PPL scores on the test set. Lower is better for both the metrics. Variance in NLL calculation is in the order of $10^{-3}$.}
\label{tab:perplex}
\end{table}

\begin{table}[h!]
\centering
\begin{tabular}{l|r|r} 
 System & NLL & PPL\\
\toprule
 SLDS-1\% & $\leq$\textbf{177.60} & \textbf{25.19} \\
 SLDS-10\% & $\leq$178.81 & 25.77 \\ 
 SLDS-25\% & $\leq$181.11 & 26.87 \\
 SLDS-50\% & $\leq$185.07 & 28.88 \\
 SLDS & $\leq$182.17 & 27.39 \\
\end{tabular}
\caption{Approximate NLL and PPL scores for SLDS and semi-supervised SLDS on the test set.}
\label{tab:eval-semis}
\end{table}

\subsection{Evaluation of Transition Dynamics}
One matter of interest is whether or not the transitions are capturing what they are supposed to capture, appropriate sentiment. 
Since we used the sentiment tagger Vader for training tags, we again utilize it to evaluate whether using transitions of a certain sentiment actually leads the model to produce outputs with the given sentiment. To perform this evaluation, we give as input to our models (and the S2S baseline) the sentiment tags for a sentence and allow it to generate a sentence conditioned on these sentiment tags. We then tag the generated sentences with Vader and see if the sentiment tags match the originals. 
We calculate the F1 score across all sentiment tags and report the macro average. In Table \ref{tab:control} we see that having labels is incredibly important for meaningful transitions. There is a large drop in F1 as the amount of labels given to the model is decreased. The SLDS model that is trained with 100\% of the labels performs a little better than even S2S, despite not having direct access to the sentiment labels (SLDS only uses the sentiment labels to decide which transition to use while the S2S model uses attention directly on the sentiment labels).

\begin{table}[h]
\centering
\begin{tabular}{l|r} 
 System & Macro F1 \\
\toprule
 S2S & 95.8 \\
 SLDS-1\% & 50.2 $\pm$ 1.1 \\
 SLDS-10\% & 51.4 $\pm$ 1.1 \\ 
 SLDS-25\% & 58.7 $\pm$ 0.4 \\
 SLDS-50\% & 74.6 $\pm$ 0.1 \\
 SLDS & \textbf{96.1 $\pm$ 0.0} \\
\end{tabular}
%\caption{how many times does the generated sentence actually fit the given sentiment tag? SLDS results averaged over 5 runs}
\caption{Macro F1 scores on sentiment classification task. Results for SLDS and SLDS-X\% are averaged over 5 runs.} 
\label{tab:control}
\end{table}

\section{Related Work}
Story/narrative generation has a rich history in the field of AI. Many early systems were based on structured formalisms for describing common narrative structures \cite{lakoff1972structural,thorndyke1977cognitive,Meehan77tale-spin}, many being inspired by the initial work of \cite{propp2010morphology}.
There has been a swath of recent work that has looked to add some semblance of a `narrative scaffold' back into generation methods \cite{Fan2018,DBLP:journals/corr/MartinAHSHR17,planandwrite,skeleton}. Many of these methods work as conditional LMs (conditioned directly on the scaffold). This line of work may be combined with our formalization as well, by conditioning the generation on the switching state as well, as done in the model of \citet{barber2006expectation}. Recent work by \citet{tambwekar2018controllable} has similar goals to ours in permitting more controlability in generation systems, developing a RL-based system that allows users to specify an end goal for a story (by specifying the event class that is desired to appear at the end). Their work differs from ours in that it does not deal with text directly, modeling only the sequences of events in the narrative. It may be possible to utilize this model as the scaffolding component in our model (utilizing their RL policy for the scaffold planner, rather than the simple Markovian distribution used here).

\section{Conclusion and Future Work}
In this paper, we formulated the problem of narrative generation as a switching dynamical system. We showed how this formulation captures notions important in narrative generation, such as narrative dynamics and scaffolds. We developed an approximate Gibbs sampling algorithm for the model that permits the system to generate interpolations conditioned on arbitrary parts of the narrative, and evaluated these interpolations using both human and automatic evaluations. Though in this work we used sentiment tags for our scaffolds/switching variables, future work may look at utilizing different kinds of information to guide the generation of narratives. Utilizing the main predicate of a sentence as a scaffold would be a logical next step, and may prove more informative then the sentiment trajectory. A scaffold such as this can take on many more possible values then a sentiment tag, and as such, it may prove difficult to assign a set of dynamics to each value. Another avenue for future work would deal with this possible problem. One potential solution could be to associate each switching variable value with a (learned) vector in a probability simplex, and use this vector to combine a small set of ``primitive" dynamics matrices in order to get that value's associated set of dynamics.

\bibliography{acl2020}
\bibliographystyle{acl_natbib}
\appendix
%auto-ignore
\section{Gibbs Sampling Derivation}
\label{sec:supplemental}
\begin{align*}
&P(Z_i | Z_{i-1}, Z_{i+1}, S_{i}, S_{i+1}, X_i, X_{i-1})  \\
&\propto P(Z_{i+1}, S_{i+1} | Z_{i-1}, Z_{i+1}, S_{i}, S_{i+1}, X_i, X_{i-1}, Z_{i}) \\
&*P(Z_{i} | Z_{i-1}, S_{i}, X_{i}, X_{i-1}) \\
&\approx  P(Z_{i+1}, S_{i+1} | Z_{i-1}, S_{i}, S_{i+1}, X_i, X_{i-1}, Z_{i}) \\
&*Q(Z_{i} | Z_{i-1}, S_{i}, X_{i}, X_{i-1}) \\
&=  P(Z_{i+1}| S_{i+1}, Z_{i-1}, S_{i}, S_{i+1}, X_i, X_{i-1}, Z_{i})\\
&*P(S_{i+1} | S_{i}) Q(Z_{i} | Z_{i-1}, S_{i}, X_{i}, X_{i-1}) \\
&\propto  P(Z_{i+1}| S_{i+1}, Z_{i-1}, S_{i}, S_{i+1}, X_i, X_{i-1}, Z_{i}) \\
&*Q(Z_{i} | Z_{i-1}, S_{i}, X_{i}, X_{i-1}) \\
&=  P(Z_{i+1}| S_{i+1},Z_{i}) Q(Z_{i} | Z_{i-1}, S_{i}, X_{i}, X_{i-1}) \\
& = \mathcal{N}_{Z_{i+1}}(A_{S_{i+1}}Z_i,\Sigma_{S_{i+1}}) \mathcal{N}_{Z_{i}}(f_{\mu}(\cdot),f_{\sigma}(\cdot))
\end{align*}
The rest can be derived by taking the PDFs of the two Gaussian densities above, getting rid of constants that don't 
depend on $Z_i$, multiplying them together, and completing the square to obtain the numerator of a Gaussian over $Z_i$ (such that $Z_i$ appears nowhere else in the equation). This numerator can then be multiplied by the normalizing constant (that does not depend on $Z_i$) to obtain exactly a Gaussian pdf with the mean and variance as given below:
\begin{align*}
\mathcal{N}_{Z_{i+1}}&(A_{S_{i+1}}Z_i,\Sigma_{S_{i+1}}) \mathcal{N}_{Z_{i}}(f_{\mu}(\cdot),f_{\sigma}(\cdot)) \\
&\propto \mathcal{N}_{Z_{i}}(\mu_*,\Sigma_*) \hphantom{1}\text{where,} \nonumber \\
\Sigma_* = &\big(A_{S_{i+1}}^T \Sigma_{S_{i+1}}^{-1} A_{S_{i+1}} + f_{\sigma}(\cdot)^{-1}\big)^{-1}\nonumber   \\
\mu_* = &\Sigma_*^T\big(Z_{i+1} \Sigma_{S_{i+1}}^{-1} A_{S_{i+1}} + f_{\mu}(\cdot)^T f_{\sigma}(\cdot)^{-1}\big)^{T}
\end{align*}

\section{Manual Error Analysis of Generations} 
We evaluate the quality of sentences of 20 generated narratives are not considered coherent by Turkers. We find that, in a broader context, 16 stories out of 20 are not good enough in terms of connecting the ending with the previous sentences. Also, in 14 out of 20 stories, \textit{mild} off-topic sentences are introduced, which are aligned with the main topic of the story but not along with local coherency (i.e. the previous and the next sentences). When considering a narrower context, or sentence level, we confirm that only 9 out of 60 generated sentences are ungrammatical, so they fail to deliver their meaning.

\section{Initializing the Gibbs Sampler}
\paragraph{Initializing $Z$}: We first initialize the $Z$ variables in the sampler. This is done as follows: The $Z$ variables are initialized in increasing order. If sentence $X_i$ is provided as input, then we sample from the approximate posterior $q_Z$ in order to initialize $Z_i$. If $X_i$ is missing, then we sample using the dynamics distribution, $P(Z_i | Z_{i-1}, S_i)$. Since we initialize in increasing order, we are guaranteed to have $Z_{i-1}$.
\paragraph{Initializing $X$}: We next initialize the missing text. Initializing the missing text $X_i$ is simply done by greedy decoding from the language model conditioned on $Z_i$, and previous sentences. 

\section{More Generated Outputs}
Below in Table~\ref{tab:sample-stories2} we provide more outputs from our system.
\begin{table*}[h]
\small
\centering
\begin{tabular}{l|l}
\toprule
\textcolor{gray}{My roomate made fish for dinner last night.} & \textbf{Mark has been working hard all week.} \\
\textbf{It was very difficult to get out of the house.} & \textbf{He calls his boss to let him know he does not know.} \\
\textbf{The next day we went to the store and complained.} & \textcolor{gray}{Finally, they gave him leave to visit his family.}\\
\textbf{The manager told me the food was terrible.} & \textcolor{gray}{He got on the plane really excited.}\\
\textcolor{gray}{I suggested he apply for the cook's position where I worked.} & \textcolor{gray}{His family were waiting for him at the airport.} \\
\midrule
\textcolor{gray}{Billy really needed help with his test.} & \textbf{Susie entered the school.} \\
\textbf{He decided to pull an all nighter.} & \textbf{She walked into the classroom.} \\
\textbf{The next day, he was ready to go to school.} & \textcolor{gray}{She explained to her teachers about her mistake.}\\
\textbf{On the day of the test he arrived.} & \textcolor{gray}{All of them believed her and said she could turn it in a day late.}\\
\textcolor{gray}{Billy failed the test.} & \textcolor{gray}{Thankfully, later that day she saw it was right where she had left it.} \\
\midrule
\textcolor{gray}{Dorothy had just turned ninety years old on Friday.} & \textcolor{gray}{Yesterday my friend and I got into a fight.} \\
\textbf{Her parents were coming to visit.} & \textbf{The argument was about to fight and we lost our temper.} \\
\textcolor{gray}{Friends and family came from all around.} & \textbf{My friend and I fought for the fight and we went to the hospital.} \\
\textbf{On the day of the trip, she hugged her parents.} & \textbf{The doctor was very nice and we were okay.} \\
\textcolor{gray}{She had a wonderful day visiting with friends and family!} & \textcolor{gray}{I'm having her over for tea later this week.} \\
\midrule
\textcolor{gray}{Sally is a writer.} & \textcolor{gray}{Tom rolled his wagon} \\
\textbf{Her writing is not great.} & \textbf{His dad was driving down the hill.} \\
\textbf{She wrote a story about her life.} & \textbf{He accidentally hit the wagon.}\\
\textbf{The story was very bad.} & \textbf{Tom's dad flipped over the handlebars.}\\
\textcolor{gray}{Sally failed in her efforts to complete the assignment.} & \textcolor{gray}{Tom's dad was not helpful ever.} \\
\bottomrule
\end{tabular}
\caption{Sample interpolations from Gibbs sampling. Grayed out lines are provided as input and bold sentences are generated by SLDS.}
\label{tab:sample-stories2}
\end{table*}

\end{document}